\documentclass[letterpaper, 10 pt, conference]{ieeeconf}  

\IEEEoverridecommandlockouts                              

\usepackage{graphics} 
\usepackage{graphicx}
\usepackage{booktabs}
\usepackage{xcolor}
\usepackage{amsmath} 
\usepackage{amssymb}  
\usepackage{hyperref}

\title{\LARGE \bf
SCOPE: Semantic Conditioning for Sim2Real Category-Level Object Pose Estimation in Robotics
}

\author{Peter Hönig$^{1}$, Stefan Thalhammer$^{2}$, Jean-Baptiste Weibel$^{3}$, Matthias Hirschmanner$^{1}$, and Markus Vincze$^{1}$ 
\thanks{*This project is funded by the FFG, project GemSort, project number FO999923008 (\href{https://www.ffg.at/}{www.ffg.at}), the Austrian
Science Fund (FWF), under project No. I 6114, project iChores, and by the EU program EC Horizon 2020 for Research and Innovation .}
\thanks{$^{1}$Peter Hönig, Matthias Hirschmanner, and Markus Vincze are with the Vision for Robotics Laboratory, Automation and Control Institute, TU Wien, Austria {\tt\small \{{hoenig, hirschmanner, vincze}\}@acin.tuwien.ac.at}}%
\thanks{$^{2}$Stefan Thalhammer is with the Department of Industrial Engineering, UAS Technikum Vienna, Vienna, Austria}
\thanks{$^{3}$Jean-Baptiste Weibel is with the Institute of Forest Engineering, BOKU University Vienna, Vienna, Austria}
}

\begin{document}

\maketitle
\thispagestyle{empty}
\pagestyle{empty}

\begin{abstract}

Object manipulation requires accurate object pose estimation. 
In open environments, robots encounter unknown objects, which requires semantic understanding in order to generalize both to known categories and beyond.
To resolve this challenge, we present SCOPE, a diffusion-based category-level object pose estimation model that eliminates the need for discrete category labels by leveraging DINOv2 features as continuous semantic priors.
By combining these DINOv2 features with photorealistic training data and a noise model for point normals, we reduce the Sim2Real gap in category-level object pose estimation.
Furthermore, injecting the continuous semantic priors via cross-attention enables SCOPE to learn canonicalized object coordinate systems across object instances beyond the distribution of known categories.
SCOPE outperforms the current state of the art in synthetically trained category-level object pose estimation, achieving a relative improvement of 31.9\% on the 5$^\circ$5cm metric.
Additional experiments on two instance-level datasets demonstrate generalization beyond known object categories, enabling grasping of unseen objects from unknown categories with a success rate of up to 100\%.
Code available: \textcolor{blue}{\url{https://github.com/hoenigpeter/scope}}.

\end{abstract}
\section{INTRODUCTION}

Autonomous manipulation and scene understanding require accurate object poses~\cite{bauer2020verefine}, with
the choice of algorithm depending on the available object priors.
If geometry priors are available, retrieving the 6D pose described by $R$ and $t$ is sufficient.
However, if the target geometry is unknown, poses need to be retrieved from similar samples encountered during training time. 
Given the requirements for robust manipulation in robotics, category-level pose estimation, where only priors regarding the semantic category or labels are known, has become the dominant approach~\cite{thalhammer2024challenges}.
Using real-world data tends to bias algorithms toward dataset specific statistics, thus harming general applicability~\cite{tobin2017domainrandomization}.
Synthetic training data is therefore preferred due to the abundance of 3D CAD models, the automatic generation of pose annotations via rendering engines~\cite{denninger2019blenderproc}, and efficient parallelization on GPUs.

\begin{figure}[t]
   \centering
    \includegraphics[width=1.0\columnwidth]{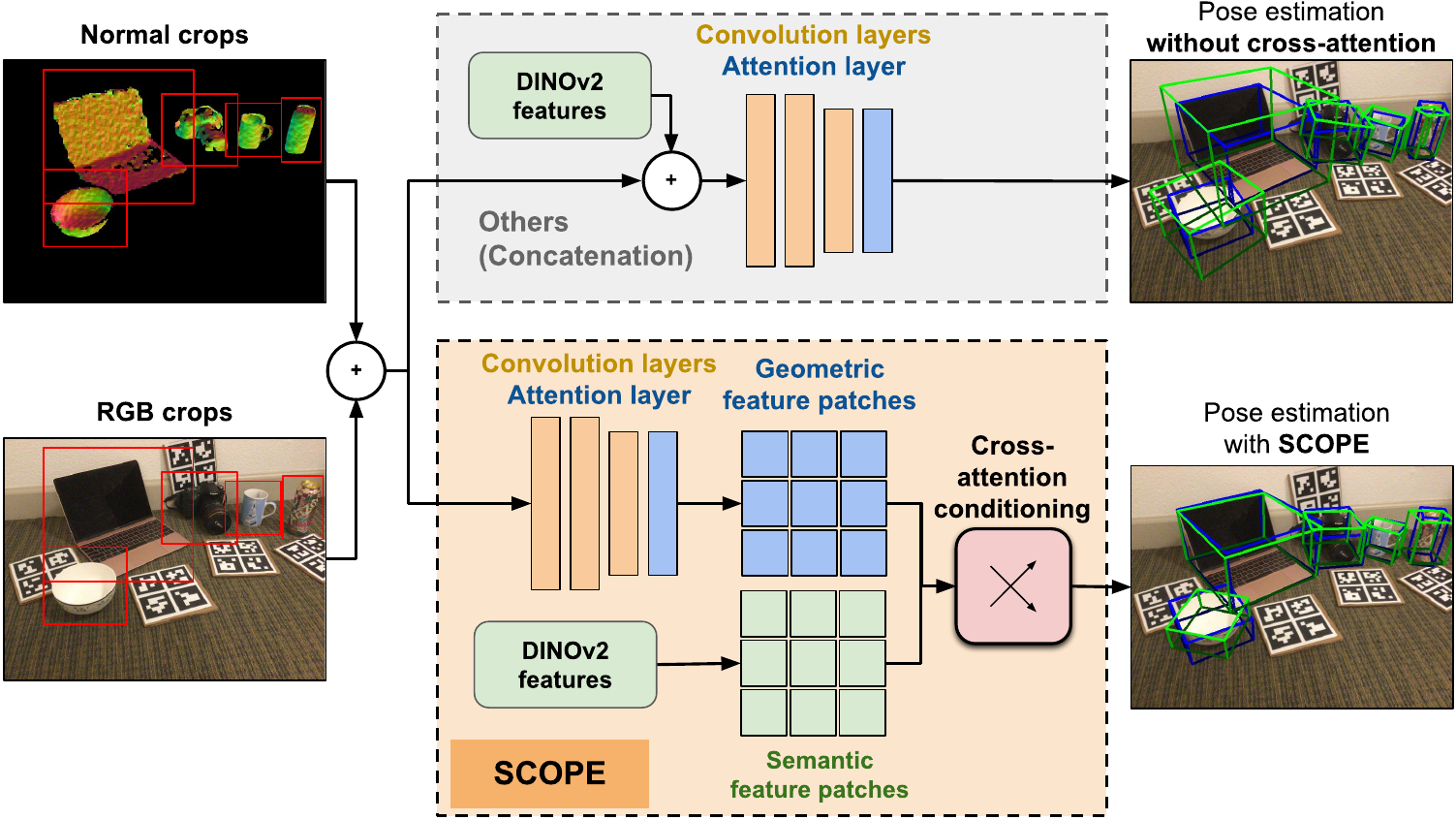}
   \caption{\textbf{SCOPE's Semantic Cross-Attention Conditioning Advantage.} Compared to prior works~\cite{ikeda2024diffusionnocs} which fuse geometric and semantic features via concatenation, we use cross-attention conditioning, significantly improving pose estimation. Ground truth poses in \textcolor{blue}{blue}, estimated poses in \textcolor{green}{green}.}
   \label{fig:improvementpic}
\end{figure}

In category-level pose estimation, using fixed category labels as priors introduces arbitrary structures into the learning process, since categories are typically defined based on human semantic conventions rather than task-relevant geometric or appearance similarities.
By contrast, learning continuous embedding spaces directly from data allows the model to organize object embeddings according to features that are more meaningful for effective generalization across instances and categories.
Consequently, learned embeddings provide a more natural and adaptive prior for category-level pose estimation than manually defined category labels~\cite{zhang2024genpose}.
Recent models~\cite{ikeda2024diffusionnocs, chen2024secondpose, wang2024gs} enhance object semantics by concatenating DINOv2~\cite{oquab2024dinov} features to the input.
This contrasts with the semantic conditioning mechanism of state-of-the-art image synthesis models such as Stable Diffusion~\cite{rombach2022high}, InstructPix2Pix~\cite{brooks2023instructpix2pix}, CogView3~\cite{zheng2024cogview3}, and FastComposer~\cite{xiao2024fastcomposer}.
These image synthesis models achieve semantic awareness through cross-attention, conditioning the generation process without requiring category or category labels.

Inspired by this semantic cross-attention conditioning, we propose a novel framework named SCOPE that injects the semantic features of DINOv2 into U-Net diffusion models via cross-attention, allowing the model to directly leverage object semantics for category-level object pose estimation and beyond.
By conditioning directly on continuous embedding spaces that describe semantics, rather than discrete category labels, our method enables pose estimation that continuously transitions between object categories and adapts to unknown instances and categories.
Fig.~\ref{fig:improvementpic} illustrates the semantic cross-attention conditioning principle of SCOPE.
Our contributions are as follows:
\begin{itemize}
    \item A category-level pose estimation method that integrates continuous DINOv2 semantics into a U-Net diffusion model via cross-attention, learning patch-level geometric-semantic correspondences without category labels, enabling category-agnostic generalization, beyond human semantic conventions.
    \item Improvements over the current state of the art on REAL275~\cite{wang2019normalized}, YCB-V~\cite{xiang2018posecnn}, and TYOL~\cite{hodan2018bop}, significantly reducing the Sim2Real gap. 
    \item Grasping experiments on the YCB-V~\cite{xiang2018posecnn} objects demonstrate that SCOPE enables grasping of previously unseen objects from known categories and generalizes to unknown categories as well.
\end{itemize}

This paper proceeds as follows: Section~\ref{sec:related_work} reviews related work.
Section~\ref{sec:method} presents our method SCOPE.
Section~\ref{sec:experiments} details the experimental setup and reports pose estimation, ablation, and grasping experiments.
We conclude in Section~\ref{sec:conclusions}.

\section{Related Work}
\label{sec:related_work}

\textbf{Category-Level Object Pose Estimation.}
Pose estimation models trained on a set of pre-defined categories, where no CAD model of the target object is available during inference, are referred to as category-level object pose estimation methods~\cite{wang2019normalized, thalhammer2024challenges}.
We limit our overview to such models that estimate the object pose from a single view.
Modern methods split the pose estimation task into multiple steps, since directly regressing the object pose from RGB or depth input has been shown to be ineffective~\cite{zhou2019continuity}. 
First, an intermediate representation~\cite{wang2019normalized, ikeda2024diffusionnocs, wang2023query6dof} such as the Normalized Object Coordinate Space (NOCS) is predicted.
The normalized point cloud is then registered with the depth data to solve for the 6D or 9D pose.
Registration between NOCS and object pose is performed using instance-based optimization algorithms such as Umeyama~\cite{umeyama1991least} or TEASER++~\cite{yang2020teaser}, or with data-driven solutions like direct regression optimization during a training phase~\cite{krishnan2024omninocs, chen2024secondpose}.

\begin{figure*}[t]
   \centering
    \includegraphics[width=1.6\columnwidth]{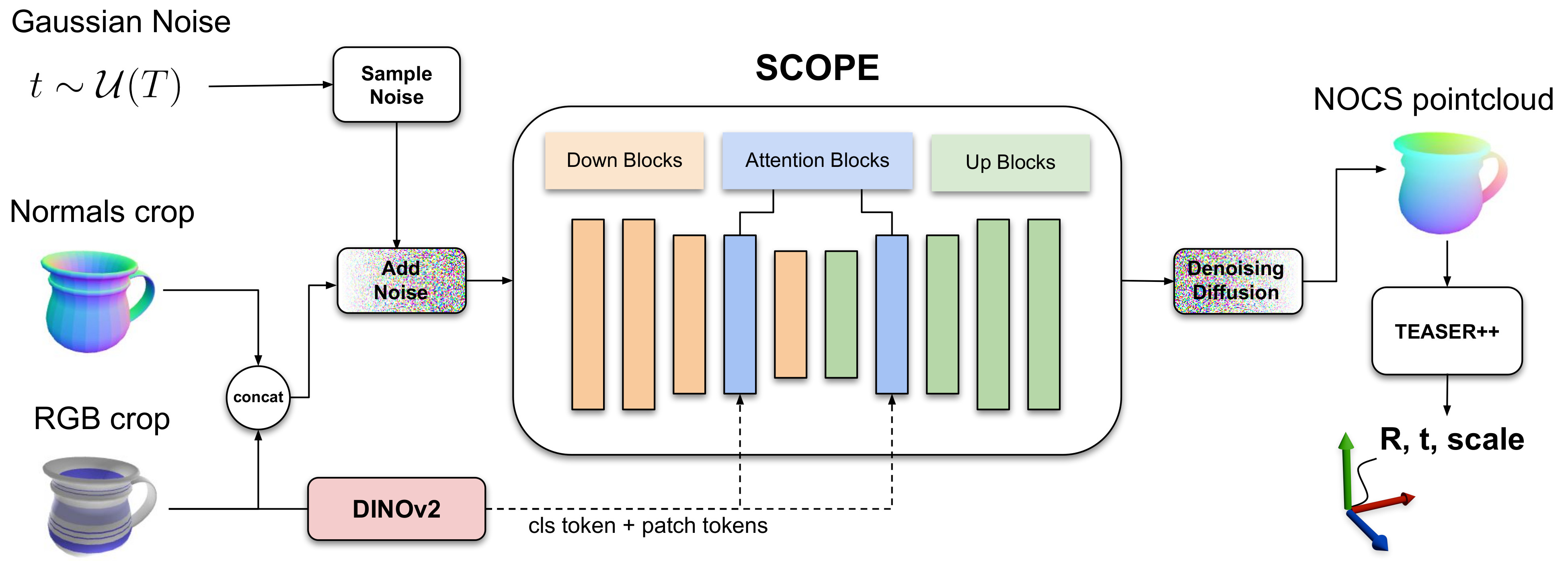}
   \caption{\textbf{Overview of SCOPE.} We concatenate normals and RGB crops at the input of a U-Net diffusion model and condition with continuous semantic DINOv2 features via cross-attention. Patch tokens and $CLS$ token of the DINOv2 embeddings provide a continuous textural, shape, and semantic feature space, aiding the diffusion model with generalization.}
   \label{fig:overview_image}
\end{figure*}

\textbf{Modalities and Conditioning.}
Since category-level object pose estimation does not assume available CAD models during inference, all shape-related information must either be acquired a priori during model optimization or during runtime from depth cues.
To aid generalization, recent methods~\cite{chen2024secondpose, ikeda2024diffusionnocs, wang2024gs} incorporate vision foundation models~\cite{oquab2024dinov}.
In~\cite{chen2024secondpose}, the authors extract DINOv2 features~\cite{oquab2024dinov} from RGB and Point-Pair Features (PPF)~\cite{drost2010ppf} from depth data.
They subsequently concatenate both feature vectors into a spherical feature space.
Similarly, in~\cite{wang2024gs} DINOv2 features are extracted and projected onto the 3D point cloud.
DINOv2 is also used in~\cite{ikeda2024diffusionnocs}, but instead of learning from explicit shape features such as PPF, features are concatenated with the surface normals and the RGB input.

While recent category-level object pose estimation methods have shown performance improvements on data limited to known categories~\cite{chen2024secondpose, zhang2024genpose}, or improved symmetry handling~\cite{ikeda2024diffusionnocs}, the transferability from known to unknown categories remains underexplored.
Similarly, recent works~\cite{chen2024secondpose, wang2024gs, ikeda2024diffusionnocs} improve the state of the art by using DINOv2 through concatenation to the model inputs.
In contrast, we aim to explore a fundamentally different approach inspired by semantic learning mechanisms in image-to-image translation models such as Stable Diffusion~\cite{rombach2022high}, InstructPix2Pix~\cite{brooks2023instructpix2pix}, CogView3~\cite{zheng2024cogview3}, and FastComposer~\cite{xiao2024fastcomposer}.
Rather than simply concatenating semantic features, we incorporate semantics through cross-attention conditioning, enabling the model to learn semantic relationships dynamically via attention.

\section{Semantic Conditioning for Object Pose Estimation (SCOPE)}
\label{sec:method}

This section presents the Semantic Conditioning for Object Pose Estimation model (SCOPE), as illustrated in Fig.~\ref{fig:overview_image}.
SCOPE is built on an attention U-Net diffusion model~\cite{oktay2018attention}, conditioned on DINOv2 features as category semantics via cross-attention.
Similar to previous works~\cite{wang2019normalized, ikeda2024diffusionnocs, tian2020shape, zhang2022ssp}, we formulate our optimization objective toward the regression of a NOCS image.
We assume a function $F$ which maps the input image $I$ and the semantic input $S$ to the estimated NOCS image $\hat{X}$, denoted by $\hat{X} = F(I, S)$.
Image input $I$ consists of the RGB image concatenated with surface normals computed from depth data, defined by \(I = \text{concat}(I_{\text{RGB}}, I_{\text{normals}}) \in \mathbb{R}^{H \times W \times 6}\).
$I_{\text{RGB}}$ and $I_{\text{normals}}$ are local objects crops, segmented from the full-scene images using the segmentation masks of an arbitrary object detector.
Semantic input $S$ is formulated as DINOv2 features containing patch and $CLS$ tokens, for zero-shot semantic grounding of unseen visual context.

We approximate function $F$ using a conditional Denoising Diffusion Probabilistic Model (DDPM)~\cite{ho2020denoising} that generates the NOCS image $X \in \mathbb{R}^{H \times W \times 3}$, which we denote as $x_0$ following DDPM notation.
During the forward pass of the DDPM model, noise $\epsilon$ is gradually added to $x_0$ over $T$ timesteps by a fixed variance schedule $\beta_t \in (0, 1)$ resulting in the noisy image $x_t$, denoted by the following equation, where \( \bar{\alpha}_t = \prod_{s=1}^t (1 - \beta_s) \) describes the cumulative noise factor up to timestep \( t \).  

\begin{equation}
x_t = \sqrt{\bar{\alpha}_t} \, x_0 + \sqrt{1 - \bar{\alpha}_t} \, \epsilon, \quad \epsilon \sim \mathcal{N}(0, \mathbf{I})
\end{equation}

To learn the reverse (denoising) process, the U-Net estimates noise \( \epsilon \) given the noisy image $x_t$, timestep \( t \), and semantic conditioning \( S \), summarized by \(\epsilon_\theta(x_t, t, I, S)\).
The training objective thereby is the minimization of the L2 loss between true and predicted noise, as shown in the following equation.

\begin{equation}
\mathcal{L}_{\text{DDPM}} = \mathbb{E}_{x_0, t, \epsilon} \left[ \left\| \epsilon - \epsilon_\theta(x_t, t, I, S) \right\|_2^2 \right]
\end{equation}

Given our estimated NOCS image $\hat{X}$, we retrieve a 6D pose and a scaling scalar, by integrating TEASER++~\cite{yang2020teaser} a fast point cloud registration algorithm built on probabilistic sampling.

\textbf{Semantic Embeddings instead of Category Priors.}
We build SCOPE on the assumption that object categories are arbitrary and do not correlate with semantics learned by machine learning algorithms~\cite{oquab2024dinov}.
While objects like a mug, a tea pot, or a pitcher exhibit semantically similar geometry and texture, they fall into different manually assigned categories.
Such cases require semantic encoding in a continuous space to reach generalization and go beyond the category-level.
To overcome this limitation, we directly employ object semantics learned by machine learning algorithms as semantic priors.
Hence, meaningful object representations are encoded, where semantically similar categories are located adjacently in the feature embedding space.

In practice, DINOv2 features are used as semantic category priors~\cite{ikeda2024diffusionnocs, chen2024secondpose, wang2024gs}.
Due to their ability to encode meaningful visual concepts, they are proven to be strong priors for zero-shot reasoning~\cite{ausserlechner2024zs6d, wang2024dfields}. 
The embedded image representation is structured such that semantically similar images produce embeddings that are proximal in the feature space and thus provide a smooth category prior transition for semantically similar concepts.

\textbf{Semantic Conditioning with Cross-Attention.}
Prior works incorporate category-specific information in two main ways, implicitly, by learning to disambiguate semantics~\cite{zhang2024genpose}, or explicitly, by concatenating semantic features to the input~\cite{ikeda2024diffusionnocs}.
This, however, differs from how semantics are propagated in diffusion models and vision transformers, which rely on the attention~\cite{vaswani2017attention} mechanism.
SCOPE instead conditions the U-Net encoder features with semantic DINOv2 embeddings via cross-attention, thereby preserving and explicitly modeling mutual spatial relationships between geometry and semantics.
This design parallels attention-based conditioning in latent diffusion models, but with a notable distinction: SCOPE learns a bidirectional correspondence between geometrical feature maps from the input (RGB + normals) and semantic patches from DINOv2.
Through cross-attention, each geometric feature map pixel can attend to every semantic patch, and vice versa, enabling the network to reason about geometric structure in the context of semantic meaning at a fine spatial granularity.
This geometric-semantic interaction makes SCOPE particularly powerful in capturing both semantic cues and fine-grained geometric details.

SCOPE employs attention layers at the fifth down-sampling block and the second up-sampling block.
The semantic input $S$ is integrated into the model via cross-attention, allowing the U-Net to attend to $S$ at each spatial location.
The cross-attention mechanism is denoted by the following equation:

\begin{equation}
\text{Attention}(Q, K, V) = \text{softmax}\left(\frac{Q K^\top}{\sqrt{d_k}}\right) V
\end{equation}

Here, \( Q \in \mathbb{R}^{H \times W \times d_k} \) describes the query, a learnable projection from the U-Net feature map at the respective attention block.
\( K, V \in \mathbb{R}^{H \times W \times d_k} \) define key and value matrices, projected from the frozen semantic embedding $S$.
$Q, K, V$ from cross-attention conditioning allow the U-Net model to attend to the semantic embeddings $S$.
In parallel, SCOPE also employs self-attention.
During the self-attention step $Q, K, V$ are all derived from the U-Net feature maps of the respective attention-layer block.
The dual-attention strategy allows SCOPE to combine textural, shape, and semantic information for enhanced shape representation, while explicitly leveraging patch-level geometric-semantic correspondences learned through cross-attention.

\textbf{Object Symmetry Disambiguation.}
While prior category-level object pose estimation methods use handcrafted symmetry annotations~\cite{wang2019normalized, zhang2022ssp, chen2021fs, tian2020shape} and symmetry-aware loss functions, our diffusion model can handle rotational symmetry implicitly, without the need for architecture changes.
This behavior of diffusion models is already demonstrated by~\cite{zhang2024genpose, ikeda2024diffusionnocs}.
The implicit symmetry handling capabilities of diffusion models stem from their foundation in partial differential equations and variational methods, which naturally preserve symmetries during the diffusion process~\cite{Alt2021DesigningRI}.

\begin{table*}[tb]
    \centering
    \caption{\textbf{Category-Level Results on REAL275.} \textbf{Bold} indicates the best method on synthetic-only training data; \underline{Underline} the best method overall.
    * ShapePrior results as reported in~\cite{ikeda2024diffusionnocs}; Gao et al.~\cite{gao2020catlevel} and Chen et al.~\cite{chen2020catlevel} results as reported in~\cite{wang2024gs}. 
    }
    \begin{tabular}{c c c | c c c | c c c c c}
        \toprule
        \textbf{Method} & \textbf{Input} & \textbf{Data} & IoU25$\uparrow$ & IoU50$\uparrow$ & IoU75$\uparrow$ & 5$^\circ$2cm$\uparrow$ & 5$^\circ$5cm$\uparrow$
        & 10$^\circ$2cm$\uparrow$ & 10$^\circ$5cm$\uparrow$ & 15$^\circ$5cm$\uparrow$ \\
        \midrule
        CenterSnap~\cite{irshad2022centersnap} & RGB-D & S+R & 83.5 & 80.2 & - & - & 27.2 & - & 58.8 & - \\
        SSP-Pose~\cite{zhang2022ssp} & RGB-D & R & 84.0 & 82.3 & 66.3 & 34.7 & 44.6 & - & 77.8 & - \\
        ShaPO~\cite{irshad2022shapo} & RGB-D & S+R & 85.3 & 79.0 & - & - & 48.8 & - & 66.8 & - \\
        DPDN~\cite{lin2022category} & RGB-D & S+R & - & 83.4 & \underline{76.0} & 46.0 & 50.7 & 70.4 & 78.4 & - \\

        GenPose~\cite{zhang2024genpose} & D & S+R & - & - & - & 52.1 & 60.9 & 72.4 & 84.0 & \underline{89.7} \\
        SecondPose~\cite{chen2024secondpose} & RGB-D & S+R & 83.7 & 66.1 & 49.7 & \underline{56.2} & \underline{63.6} & \underline{74.7} & \underline{86.0} & - \\
        \midrule
        Chen et al.~\cite{chen2020catlevel}* & RGB-D & S & 15.5 & 1.3 & - & - & 0.7 & - & 3.6 & 9.1 \\
        Gao et al.~\cite{gao2020catlevel}* & D & S & 68.6 & 24.7 & - & - & 7.8 & -& 17.1 & 26.5 \\
        ShapePrior~\cite{tian2020shape}* & RGB-D & S & - & - & - & - & 12.0 & - & 37.9 & 52.8 \\

        CPPF~\cite{you2022cppf} & D & S & 78.2 & 26.4 & - & - & 16.9 & - &  44.9 & 50.8 \\

       i2c-Net~\cite{remus2023icnet} & RGB-D & S & \underline{\textbf{99.9}} & \underline{\textbf{92.5}} & - & - & 24.6 & - & 49.4 & - \\

        GS-Pose~\cite{wang2024gs} & RGB-D & S & 82.1 & 63.2 & - & - & 28.8 & - & 60.1 & 73.6 \\
        DiffusionNOCS~\cite{ikeda2024diffusionnocs} & RGB-D & S & - & - & - & - & 35.0 & - & 66.6 & 77.1 \\
        DPDN~\cite{lin2022category} & RGB-D & S & 71.7 & 60.8 & - & 29.7 & 37.3 & 53.7 & 67.0 & - \\
        \midrule
        
        \textbf{SCOPE (CAMERA)} & RGB-D & S & 83.7 & 83.0 & \textbf{73.7} & \textbf{42.6} & \textbf{49.2} & \textbf{63.4} & 73.4 & 79.1 \\
        \textbf{SCOPE (CAMERA-BPR)} & RGB-D & S & 84.1 & 82.8 & 61.6 & 42.2 & 49.0 & 63.3 & \textbf{73.8} & \textbf{83.4} \\
        \bottomrule
    \end{tabular}
    \label{tab:pose_estimation}
\end{table*}

\section{Experiments}
\label{sec:experiments}
We structure our experiments as follows. We first describe our experimental setup, then address three core questions: (Section~\ref{sec:catlevel_pose_estimation_performance}) How does SCOPE perform on category-level object pose estimation compared to state-of-the-art models trained on synthetic or mixed data? (Section~\ref{sec:instance_level_pose_estimation_performance}) Can SCOPE generalize to pose estimation beyond known categories? (Section~\ref{sec:grasping_experiments}) Is SCOPE capable of enabling robotic manipulation of previously unseen objects from known and unknown categories?

\subsection{Experimental Setup}
\label{sec:experimental_setup}
SCOPE's U-Net~\cite{oktay2018attention} uses an image size of $height = width = 160$, to allow for direct comparison with~\cite{ikeda2024diffusionnocs}.
Denoising is implemented using a DDPM~\cite{ho2020denoising}, trained for 50 epochs, with 1000 denoising timesteps.
The Adam optimizer is parameterized with a learning rate of $0.0001$, $\beta_1=0.9$, $\beta_2=0.999$, and an $\epsilon=1e-8$ for numerical stability.
The learning rate is gradually increased with a warm-up phase of 1000 steps.
During runtime we use the fast DPM++ solver~\cite{lu2022dpmsolver} with 5 denoising steps, if not otherwise specified.
For TEASER++, we define the noise bound with $0.02$, the maximum iteration number for rotation estimation with $1000$, and a rotation cost threshold of $1e-12$.


\textbf{Synthetic Data Generation.}
All reported experiments are conducted on the 6 categories of CAMERA and REAL275~\cite{wang2019normalized}: Bottle, bowl, camera, can, laptop, and mug.
We train on two different datasets.
The synthetic CAMERA~\cite{wang2019normalized} dataset is used to ensure a fair comparison to the state of the art in synthetically and mixed-trained category-level object pose estimation on the REAL275 benchmark.
However, the CAMERA dataset exhibits a strong bias toward upright object poses and low occlusion.
To address this, we render a second variant of the CAMERA dataset, named CAMERA-BlenderProc Randomized (CAMERA-BPR), following the standard domain randomization~\cite{tobin2017domainrandomization} procedure employed by the Benchmark on Object Pose Estimation (BOP) datasets~\cite{sundermeyer2023bop}, using BlenderProc~\cite{denninger2019blenderproc}.
Object rotations are fully randomized.
Each scene contains two random objects per category, totaling 12 objects per scene across six categories. We render 25 camera views per scene for 2,000 scenes, resulting in 50,000 images (a fraction of the 300,000 images in CAMERA) and 600,000 object instances overall.

We apply the following online data augmentations to both CAMERA variants:
Depth data is augmented using Perlin noise, an effective augmentation type for reducing the Sim2Real domain gap~\cite{thalhammer2019sydpose}.
RGB and normals with a 2D visibility below $50\%$ are cropped based on rendered segmentation masks.
Normals are augmented with random rectangular dropout (probability 50\%, 5\% of crop size) and pixel-wise dropout (between 0 and 10\% of pixels set to 0).

\begin{table*}[tb]
\centering
\caption{\textbf{Category-Level Results on the YCB-V and TYOL.} \textbf{Bold} for best method on synthetic-only training data; \underline{Underline} for best method overall. Evaluation on the bottle, bowl, can, and mug objects. Numbers for comparison from~\cite{ikeda2024diffusionnocs}.}
\begin{tabular}{l|c|ccc|ccc}
\toprule
\textbf{Method} & \textbf{Data} & \multicolumn{3}{c|}{\textbf{YCB-V}} & \multicolumn{3}{c}{\textbf{TYOL}} \\
 &  & 5°5cm$\uparrow$ & 10°5cm$\uparrow$ & 15°5cm$\uparrow$ & 5°5cm$\uparrow$ & 10°5cm$\uparrow$ & 15°5cm$\uparrow$ \\
 \midrule
 ShapePrior~\cite{tian2020shape} & S+R & 27.4 & 54.3 & 66.9 & 14.9 & 27.5 & 32.1 \\
 DualPoseNet~\cite{lin2021dualposenet} & S+R & 36.6 & \underline{78.8} & \underline{87.5} & 22.5 & 28.8 & 32.2 \\
 GenPose~\cite{zhang2024genpose} & S+R & \underline{48.1} & 68.8 & 73.8 & 20.1 & 23.7 & 24.7 \\
  \midrule
CPPF~\cite{you2022cppf} & S & 8.8 & 35.2 & 46.9 & 13.2 & 18.8 & 20.8 \\
ShapePrior~\cite{tian2020shape} & S & 11.0 & 40.3 & 49.1 & 17.3 & 26.5 & 33.4 \\
DiffusionNOCS~\cite{ikeda2024diffusionnocs} & S & 23.6 & 53.7 & 56.6 & 43.7 & 59.0 & 64.7 \\
\midrule
\textbf{SCOPE (CAMERA)} & S & \textbf{47.6} & \textbf{71.6} & \textbf{76.8} & 21.7 & 24.3 & 24.9 \\
\textbf{SCOPE (CAMERA-BPR)} & S & 40.5 & 61.6 & 75.4 & \underline{\textbf{56.6}} & \underline{\textbf{73.3}} & \underline{\textbf{77.5}} \\
\bottomrule
\end{tabular}
\label{tab:ycbv_tyol_catlevel_results}
\end{table*}

\begin{figure*}[tb]
   \centering
   \includegraphics[width=1.5\columnwidth]{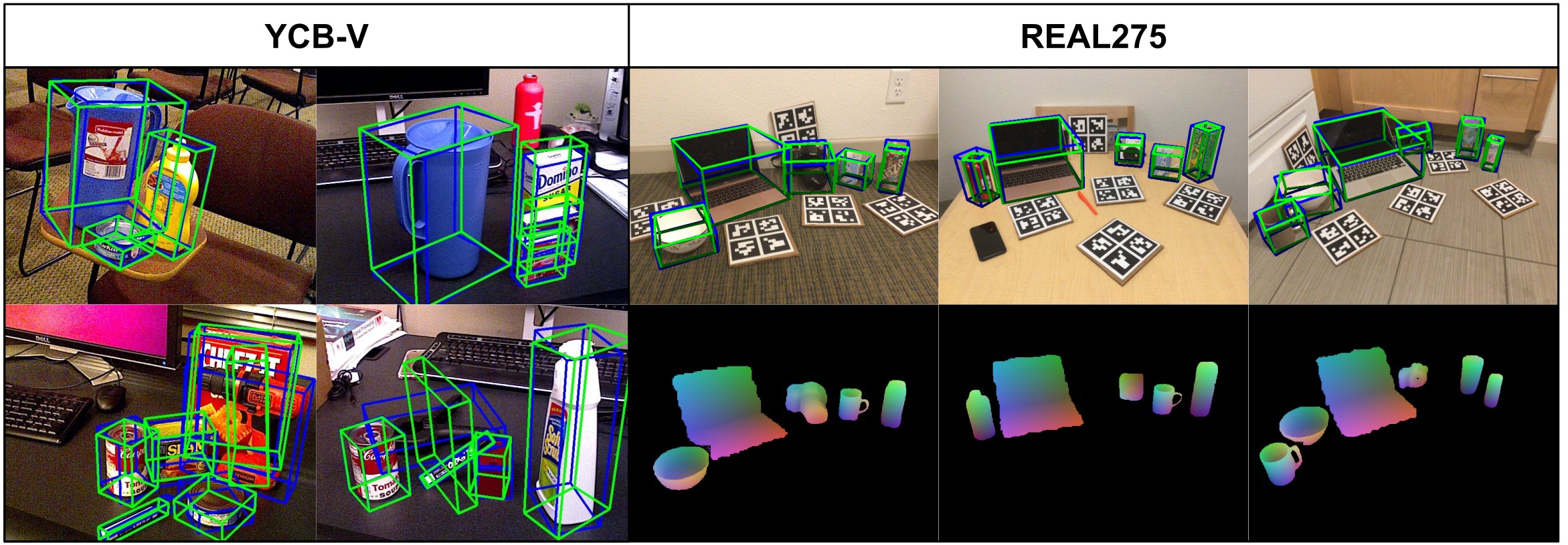}
   \caption{\textbf{Exemplary Images of SCOPE on YCB-V and REAL275 Test Images.} Ground truth pose in \textcolor{blue}{blue}, estimated pose in \textcolor{green}{green}; estimated NOCS images for REAL275.}
   \label{fig:comparison}
\end{figure*}

\textbf{Evaluation Metrics.}
To ensure comparability to the category-level state of the art on REAL275, YCB-V, and TYOL, we use the average recall of rotational and translational error of estimated poses, as well as the Intersection over Union (IoU) recalls at given thresholds, following~\cite{wang2019normalized, chen2020catlevel, lin2021dualposenet, irshad2022shapo, zhang2024genpose}.
We report standard error thresholds of 5$^\circ$5cm, 10$^\circ$5cm, and 15$^\circ$5cm throughout all experiments, and additionally report 25\%, 50\%, and 75\% IoU recalls, as well as 5$^\circ$2cm and 10$^\circ$2cm specifically for the REAL275 benchmark.
For instance-level analysis on YCB-V and TYOL we adopt the evaluation metrics from~\cite{hinterstoisser2012model}, namely the ADD(-S) and ADD-S scores with a threshold of 10\% of the model diameter.
However, since textural cues are treated as unknown in category-level pose estimation we only consider shape as a symmetry criterion.

\subsection{Category-Level Pose Estimation Results}
\label{sec:catlevel_pose_estimation_performance}

Table \ref{tab:pose_estimation} compares SCOPE to the state of the art for category-level pose estimation on REAL275.
The two SCOPE variants exhibit similar performance across most metrics, with the exception of IoU75, where the CAMERA-PBR variant shows a performance decrease.
This drop is likely due to the stronger domain adaptation of the CAMERA variant toward the specific characteristics of REAL275.
To ensure comparability, we use the same pre-computed 2D bounding box and segmentation priors as all other methods listed in Table~\ref{tab:pose_estimation}.

Regarding 3D bounding box accuracy, i2c-Net~\cite{remus2023icnet} reports higher IoU25 and IoU50 scores, but does not provide IoU75 results.
Among methods that report all three IoU thresholds, SCOPE (CAMERA) achieves the highest IoU75, outperforming the previous synthetic-only benchmark leader GS-Pose~\cite{wang2024gs} by 16.5\% at IoU50 and 31.3\% at IoU75, while almost matching the performance of the mixed-data benchmark leader DPDN~\cite{lin2022category} at IoU50.

In terms of rotation and translation estimation accuracy, SCOPE surpasses the leading synthetic benchmark method DPDN~\cite{lin2022category} by 31.9\% and 9.6\% on the 5$^\circ$5cm and 10$^\circ$5cm metrics, respectively.
Compared to methods trained on a combination of real and synthetic data, SCOPE outperforms ShaPO~\cite{irshad2022shapo}, SSP-Pose~\cite{zhang2022ssp} and CenterSnap~\cite{irshad2022centersnap}.
However, accuracy of SCOPE falls short relative to GenPose~\cite{zhang2024genpose} and SecondPose~\cite{chen2024secondpose}, which benefit from access to real data in the target domain, including depth noise characteristics and scene properties.
Notably, SCOPE improves considerably over the mixed data trained SecondPose regarding 3D bounding box accuracy with 25.6\% and 48.3\% on the IoU50 and IoU75 metrics, respectively.
Qualitative results on REAL275, including estimated NOCS images and 3D bounding boxes derived from predicted 6D poses and object scales, are presented in Fig.~\ref{fig:comparison}.

Table~\ref{tab:ycbv_tyol_catlevel_results} presents category-level results on the YCB-V and TYOL datasets for known categories (bottle, bowl, can, mug).
While the performance of both SCOPE variants is similar on YCB-V, the domain randomization in CAMERA-PBR improves performance compared to the CAMERA variant on the TYOL dataset, which contains both upright and flat-lying objects.
SCOPE (CAMERA-PBR) surpasses the state of the art across both synthetic-only and mixed (synthetic + real) training data on TYOL.
On YCB-V, both SCOPE variants surpass the state of the art for synthetic-only models.
The Sim2Real gap between SCOPE and the mixed-data trained benchmarks is reduced to 1.1\% on the most precise 5$^\circ$5cm metric.

While DINOv2 has become the standard for semantic feature extraction in category-level object pose estimation~\cite{chen2024secondpose, ikeda2024diffusionnocs, wang2024gs}, CLIP~\cite{lin2024clipose} remains a widely adopted alternative.
To assess the impact of the feature backbone, we compare both in Table~\ref{tab:clip_vs_dinov2}.
The results show a performance drop of the CLIP-conditioned variant, which confirms that CLIP excels at text-image alignment, but DINOv2 offers stronger spatial consistency and fine-grained object representations, which are essential for pixel-aligned guidance for NOCS regression.

\begin{table}[tb]
\centering
\caption{\textbf{CLIP vs. DINOv2 Features.} Pose accuracy (\%) at different thresholds. CAMERA-PBR training data.}
\begin{tabular}{lccc}
\toprule
\textbf{Feature Extractor} & 5$^\circ$5cm$\uparrow$ & 10$^\circ$5cm$\uparrow$ & 15$^\circ$5cm$\uparrow$ \\
\midrule
CLIP-ViT/L14 & 28.1 & 52.8 & 59.9 \\
DINOv2 & \textbf{49.0} & \textbf{73.8} & \textbf{83.4} \\
\bottomrule
\end{tabular}
\label{tab:clip_vs_dinov2}
\end{table}

\begin{table}[tb]
\centering
\caption{\textbf{Beyond Category-Level Results.} ADD-S, ADD(-S) (\%) on YCB-V and TYOL test sets. CAMERA-PBR training data.}
\begin{tabular}{l rr rr}
\toprule
\textbf{Method} & \multicolumn{2}{c}{\textbf{YCB-V}} & \multicolumn{2}{c}{\textbf{TYOL}} \\
 & ADD-S & ADD(-S) & ADD-S & ADD(-S) \\
\midrule
Concat. + category label & 59.5 & 54.5 & 60.6 & 42.3 \\
Cross-attn. conditioning & \textbf{76.0} & \textbf{67.1} & \textbf{82.5} & \textbf{70.2} \\
\bottomrule
\end{tabular}
\label{tab:scope_vs_diffusionnocs}
\end{table}

\begin{table*}[tb]
    \centering
    \caption{\textbf{Results of Ablation Studies.} Ablation of sampled points (5 denoising steps, 1 refinement step), denoising steps (500 points for sampling, 1 refinement step), refinement steps (500 points for sampling, 5 denoising steps), and runtime analysis. Runtime results for NOCS (preprocessing + denoising diffusion) and TEASER++ (pointcloud registration + postprocessing) on a RTX 3090 GPU and Ryzen 9 5900X CPU.}
    \begin{tabular}{l|ccccc|ccccc|cccc}
        \toprule
        \textbf{Metric} & \multicolumn{5}{c|}{\textbf{Sampled Points}} & \multicolumn{5}{c|}{\textbf{Denoising Steps}} & \multicolumn{4}{c}{\textbf{Refinement Steps}} \\
        \midrule
        & 50 & 100 & 200 & 500 & 1000 & 1 & 3 & 5 & 7 & 10 & 1 & 3 & 6 & 10 \\
        \midrule
        IoU75$\uparrow$ & 70.2 & 72.1 & 72.6 & \textbf{73.7} & 73.6 &  9.0 & 72.9 & \textbf{73.7} & 73.0 & 73.1 & 73.3 & 74.1 & 74.4 & \textbf{74.6} \\
        5$^\circ$5cm$\uparrow$  & 45.5 & 47.9 & 49.0 & 49.2 & \textbf{50.3} &  5.5 & 46.0 & 49.2 & \textbf{49.8} & 49.4 & 49.2 & 50.0 & 50.1 & \textbf{50.4} \\
        10$^\circ$5cm$\uparrow$ & 72.2 & 72.8 & 73.2 & \textbf{73.4} & 73.1 & 24.5 & 71.9 & 73.3 & 73.5 & \textbf{73.7} & 73.1 & 73.7 & \textbf{74.0} & 73.9 \\
        15$^\circ$5cm$\uparrow$ & 79.2 & 79.2 & \textbf{79.4} & 79.1 & 79.2 & 36.3 & 78.4 & 79.3 & 79.3 & \textbf{79.5} & 79.2 & 79.8 & 79.9 & \textbf{80.1} \\
        \midrule
        Time NOCS (s) & 0.133 & 0.133 & 0.133 & 0.133 & 0.133 & 0.059 & 0.094 & 0.133 & 0.162 & 0.220 & 0.133 & 0.315 & 0.621 & 1.010 \\
        Time TEASER (s) & 0.021 & 0.022 & 0.026 & 0.060 & 0.240 & 0.060 & 0.060 & 0.060 & 0.060 & 0.060 & 0.060 & 0.173 & 0.360 & 0.622 \\
        Time Total (s) & \textbf{0.154} & 0.155 & 0.159 & 0.193 & 0.373 & \textbf{0.119} & 0.154 & 0.193 & 0.222 & 0.280 & \textbf{0.193} & 0.488 & 0.981 & 1.632\\
        \bottomrule
    \end{tabular}
    \label{tab:ablations}
\end{table*}

\subsection{Beyond Category-Level Pose Estimation Results}
\label{sec:instance_level_pose_estimation_performance}

We further investigate the generalization capability of SCOPE by evaluating pose estimation on all objects from YCB-V and TYOL.
We compare the label-free cross-attention conditioning configuration of SCOPE with a configuration using concatenation of DINOv2 features to the input (RGB + normals) and discrete category labels to control semantics.
For the concatenation, we use DINOv2 features and category label conditioning identical to DiffusionNOCS~\cite{ikeda2024diffusionnocs}.
To assign category labels to the objects of YCB-V and TYOL which do not fit the known categories of CAMERA, we prompt ChatGPT~\cite{openai2023chatgpt} with: \textit{``assign the most similar category to the given instance name, given a set of possible category names".}
This category labeling is solely necessary for the concatenation example and not for the default variant of SCOPE.

Table~\ref{tab:scope_vs_diffusionnocs} reports the results of SCOPE on YCB-V and TYOL, considering all 21 objects in YCB-V and all 21 objects in TYOL.
On both datasets, the cross-attention conditioning variant of SCOPE consistently outperforms the concatenation and category label variant across the two evaluated metrics ADD-S and ADD(-S).
On YCB-V, SCOPE achieves relative improvements of 27.7\% in ADD-S and 23.1\% in ADD(-S).
Similarly, on TYOL, SCOPE shows performance gains of 36.1\% and 66.0\% respectively.
The performance gap is larger on TYOL, suggesting improved semantic awareness and stronger generalization capabilities of SCOPE's cross-attention conditioning.

\begin{figure*}[tb]
   \centering
    \includegraphics[width=1.6\columnwidth]{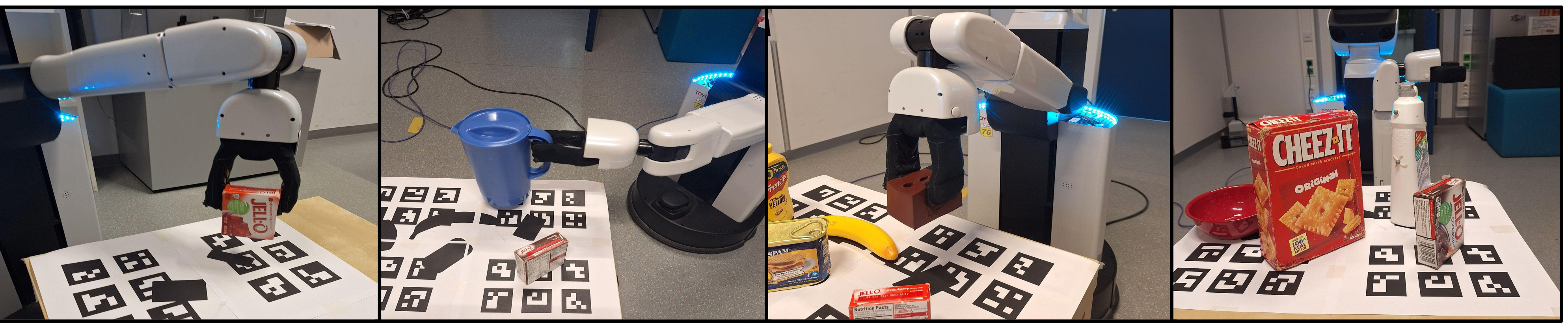}
   \caption{\textbf{Exemplary Images from Grasp Experiments.} SCOPE enables successful grasps of unseen objects from known and unknown categories.}
   \label{fig:grasping_images}
\end{figure*}

\subsection{Ablations}
\label{sec:ablations}


\textbf{Number of Sampled Points.}
Runtime depends on the number of sampled points for TEASER++ registration.
As shown in Table~\ref{tab:ablations}, maximum performance on the IoU75 and 10$^\circ$5cm metrics is achieved using 500 sampled points.
Fewer points result in negligible performance loss while gaining improvement regarding runtime.

\textbf{Number of Denoising Steps.}
Results for denoising steps ablation are presented in Table~\ref{tab:ablations}.
The authors of the DPM++ solver~\cite{lu2022dpmsolver} report ideal results at 15-20 denoising steps.
Starting our ablation at 1 inference step, increasing to 3, 5, 7, and 10, we find that 10 denoising steps yield maximum performance on the 10$^\circ$5cm and 15$^\circ$5cm metrics, while 5 denoising steps yield maximum performance on IoU75.
A balanced parameter setup between speed and accuracy with 5 denoising steps, 500 sampled points and no refinement results in an average runtime of 0.193\,s.

\textbf{Number of Refinement Steps.}
Since SCOPE uses a denoising diffusion model, noise has to be sampled for each inference run from a Gaussian distribution.
Refinement therefore refers to multiple noise samples for the same image and semantic input, similar to the refinement strategy of~\cite{ikeda2024diffusionnocs}.
As shown in Table~\ref{tab:ablations}, this refinement strategy has less influence on SCOPE, with a relative performance increase of 2.4\% on the 5$^\circ$5cm metric but an increase in runtime of 746\%.
Therefore, we define the optimal hyperparameter combination as 5 denoising steps, 500 sampled points and a single inference step for fast runtime of 0.193\,s and negligible performance loss.

\subsection{Grasping Experiments}
\label{sec:grasping_experiments}
This section demonstrates SCOPE's capability of manipulating known and unknown object categories.
We use the SCOPE model trained on CAMERA-BPR.
Experiments are conducted with the Toyota HSR robot~\cite{yamamoto2018development} using the YCB-V objects.
The YCB-V dataset features objects from known categories (bottle, bowl, can, mug), as well as objects from unknown categories (multiple box-shaped objects, foam brick, banana, scissors, pitcher).
To ensure a fair experimental setup, we use the GRASPA~\cite{bottarel2020graspa} object layout.
For objects not featured in GRASPA, we use the footprints of the featured objects to ensure clutter with canonicalized rotations.

\begin{figure}[tb]
   \centering
    \includegraphics[width=1.0\columnwidth]{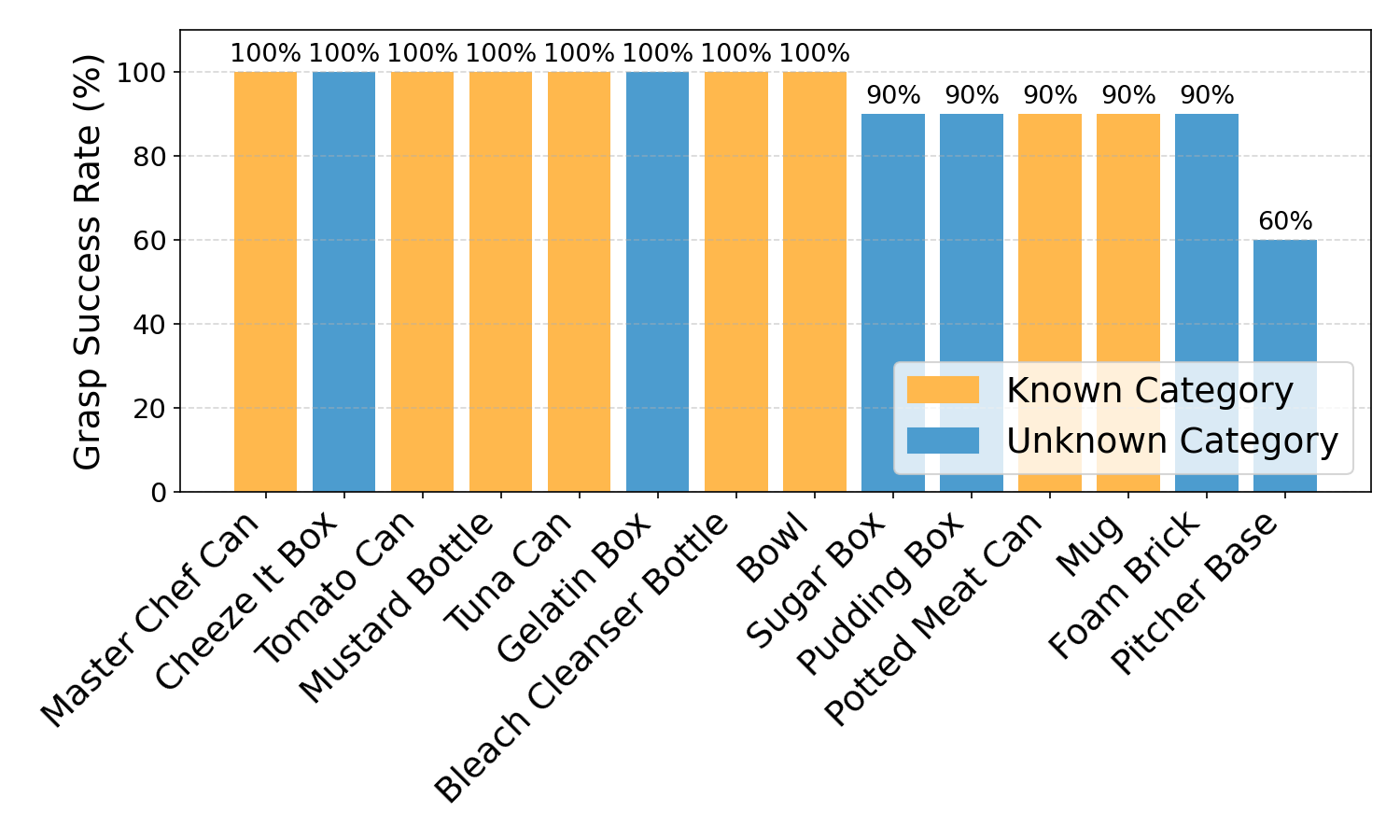}
    \caption{\textbf{Grasp Success Rates (SR) (\%) for YCB-V Objects.} Banana, marker pen, clamps, and wood block were excluded due to being too flat to grasp or exceeding the maximum robot payload.}
   \label{fig:grasping_results}
\end{figure}

Since we want to isolate our investigation to the quality of rotation and translation estimation and do not estimate grasping points, we hand-annotate grasps for each object via Blender using the HSR gripper and YCB-V CAD models.
Object detection and segmentation are performed using YOLOv8n~\cite{Jocher_Ultralytics_YOLO_2023} (75 FPS on an RTX 3090 GPU and Ryzen 9 5900X CPU).
For each detected object, SCOPE estimates poses, and grasp hypotheses are checked against the robot's physical constraints and the surrounding collision environment.
The final trajectory of the robot is calculated with MoveIt~\cite{moveit}.
Success rate refers to the number of stable grasps executed by the robot, divided by the number of all grasping attempts.
Ten grasping attempts per object are conducted.
Fig.~\ref{fig:grasping_results} shows the results of the grasping experiments, while Fig.~\ref{fig:grasping_images} shows exemplary images of the grasping experiments. 

SCOPE enables a 90-100\% grasp success rate for box-shaped objects of various sizes, including the foam brick, despite boxes being outside the known category distribution.
The lowest success rate is 60\% for the unknown object pitcher base.
Grasping attempts failed for the pitcher base due to the robot missing the handle, where only slight rotation errors can result in failure.
SCOPE also successfully deals with strong intra-category variation, leading to grasp success rates between 90-100\% for known object categories.
Our grasping experiments indicate that SCOPE is not only beneficial for Sim2Real category-level object pose estimation, but also demonstrates promising results toward generalization to object instances outside known categories enabling usage in real-world robotic scenarios.

\section{Conclusions}
\label{sec:conclusions}
We present Semantic Conditioning for Object Pose Estimation (SCOPE) for Sim2Real category-level object pose estimation.
SCOPE estimates the 6D object pose and scale from any given segmented RGB and normals crop, by leveraging semantic DINOv2 features through cross-attention conditioning of a denoising diffusion U-Net model.
This mechanism of using continuous semantic priors enables SCOPE to improve performance on unknown instances and manipulate unseen objects beyond known categories.
Using the six categories from CAMERA and REAL275, our experiments reduce the Sim2Real gap and show promising performance for pose estimation on YCB-V and TYOL.
Further research will focus on improving SCOPE by increasing the training category diversity regarding semantics and geometry to achieve universal pose estimation.
We will also experiment with cross-attention conditioning for geometrical features from using foundational models on point clouds or meshes.

{\small
\bibliographystyle{IEEEtranS}
\bibliography{bib}
}

\end{document}